\documentclass[table,runningheads]{llncs}
\usepackage[colorlinks]{hyperref}
\usepackage[T1]{fontenc}
\usepackage{graphicx,verbatim}
\usepackage{amsmath} 
\usepackage{cleveref}
\usepackage{todonotes}
\usepackage{multirow}

\definecolor{g}{RGB}{189, 230, 205}
\definecolor{l}{RGB}{228,238,188}          
\definecolor{ll}{RGB}{255,248,197}

\definecolor{o}{RGB}{228,238,188}
\definecolor{r}{RGB}{255, 226, 212} 
\definecolor{lll}{RGB}{255,248,227} 
\usepackage{pifont}
\usepackage{marvosym}

\usepackage[marginal]{footmisc}

\usepackage{xcolor}
\begin{document}

\title{Pre-Trained LLM is a Semantic-Aware and Generalizable Segmentation Booster}
\titlerunning{LLM4Seg}
%

\author{
Fenghe Tang\inst{1,2}$^\dagger$ 
\and Wenxin Ma\inst{1,2}$^\dagger$  
\and Zhiyang He\inst{3}  
\and \\ Xiaodong Tao\inst{3}  
\and Zihang Jiang\inst{1,2} $^{\href{mailto:jzh0103@ustc.edu.cn}{\textrm{\Letter}}}$  
\and S. Kevin Zhou\inst{1,2} 
$^{\href{mailto:skevinzhou@ustc.edu.cn}{\textrm{\Letter}}}$} 

\authorrunning{Fenghe Tang, Wenxin Ma, et al.}

\institute{School of Biomedical Engineering, Division of Life Sciences and Medicine, University of Science and Technology of China, Hefei, Anhui, 230026, P.R. China \and
Center for Medical Imaging, Robotics, Analytic Computing \& Learning (MIRACLE), Suzhou Institute for Advance Research, USTC, 215123, P.R. China \and
Anhui IFLYTEK CO., Ltd. \\
\email{fhtan9@mail.ustc.edu.cn, wxma@mail.ustc.edu.cn}\\
\url{https://github.com/FengheTan9/LLM4Seg}}



\maketitle              
\footnote{\textsuperscript{$\dagger$}Equal contribution. \textsuperscript{\Letter}Corresponding author.}

\begin{figure}[!]
    \vspace{-15mm}
    \centering
    \includegraphics[width=0.92\linewidth]{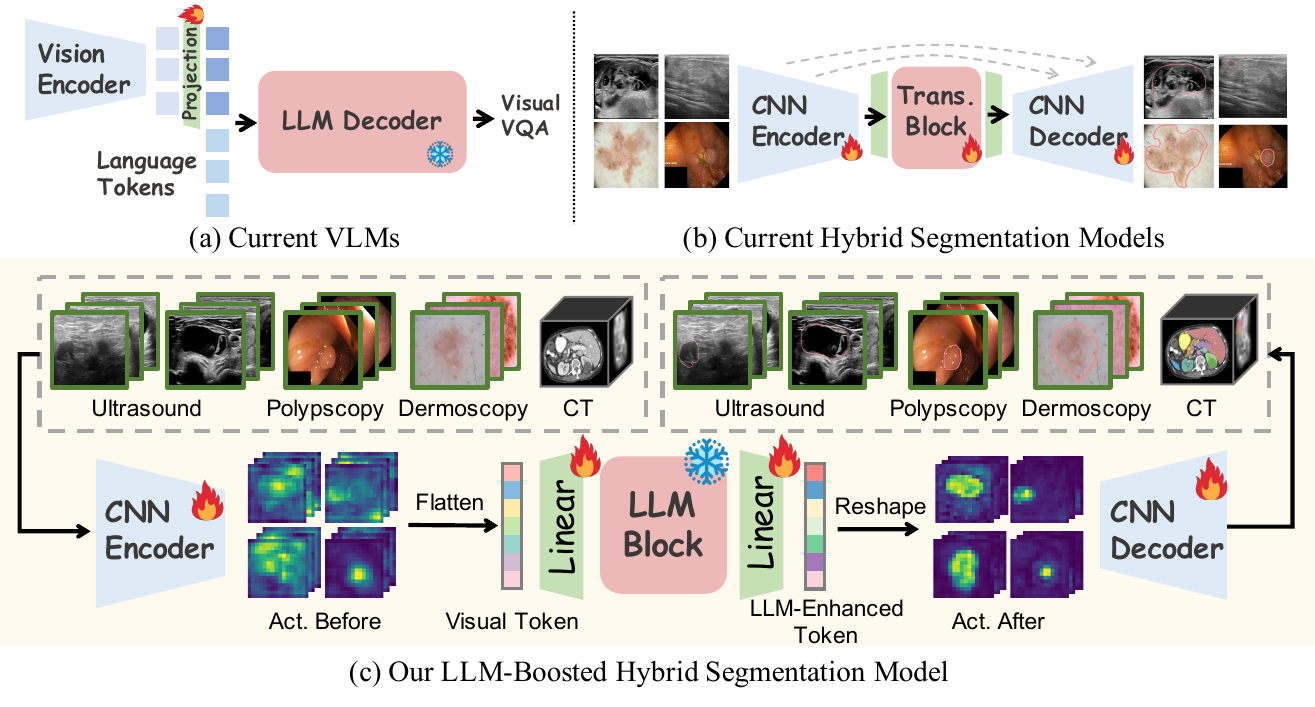}
    \vspace{-4mm}
    \caption{Comparison of current VLMs, hybrid segmentation models, and our novel hybrid segmentation model. Our LLM4Seg use a frozen LLM layer within CNN encoder-decoder framework to boost global visual understanding.}
    \vspace{-14mm}
    \label{fig:teaser}
\end{figure}

\begin{abstract}
With the advancement of Large Language Model (LLM) for natural language processing, this paper presents an intriguing finding: a frozen pre-trained LLM layer can process visual tokens for medical image segmentation tasks. Specifically, we propose a simple hybrid  structure that integrates a pre-trained, frozen LLM layer within the CNN encoder-decoder segmentation framework (LLM4Seg). Surprisingly, this design improves segmentation performance with a minimal increase in trainable parameters across various modalities, including ultrasound, dermoscopy, polypscopy, and CT scans. Our in-depth analysis reveals the potential of transferring LLM's semantic awareness to enhance segmentation tasks, offering both improved global understanding and better local modeling capabilities. The improvement proves robust across different LLMs, validated using LLaMA and DeepSeek.
\vspace{-2mm}

\end{abstract}
\section{Introduction}
Medical image segmentation is crucial for diagnosis, surgical planning, and disease monitoring. While Convolutional Neural Networks (CNNs) effectively capture local features, Vision Transformers (ViTs) have gained attention for their strong global semantic learning capabilities in segmentation tasks~\cite{cheng2022masked,lin2022ds,cao2022swin,hiend,unetr}, driving the development of hybrid structures that leverage their complementary strengths to achieve better performance than pure CNN-/transformer-based methods~\cite{chen2021transunet,tang2023mobileutr,cao2022swin,hyspark,mambamim}. However, it requires large-scale training data, otherwise it can fall in local shortcuts and struggle to learn robust semantic representations, posing a significant challenge in label-scarcity medical tasks~\cite{dosovitskiy2020image}.  

Notably, Large Language Models (LLMs), built upon transformer blocks and trained on massive linguistic datasets, have revolutionized natural language processing (NLP) with their advanced reasoning, contextual understanding, and generalization capabilities~\cite{llama3,deepseek}. While originally designed for NLP, recent multi-modal researches have explored LLM's interaction with visual representations, developping various Visual Language Models (VLMs)~\cite{li2024llava,li2023blip,pang2023frozen}. Specifically, these methods adapt visual representations to the language space, as shown in Fig.~\ref{fig:teaser}(a), which typically use an LLM as a versatile decoder for both linguistic and adapted visual tokens. Surprisingly, as validated in~\cite{pang2023frozen}, LLMs exhibit remarkable semantic understanding of visual tokens with strong generalizability. This raises an intriguing question as we investigate their potential in medical segmentation: \textit{Is it possible to directly use pre-trained LLMs in hybrid structures to process visual tokens for medical image segmentation?}

To validate the feasibility, unlike conventional multi-modal methods that use LLMs as a shared decoder, we innovatively propose a novel hybrid model with simple yet effective hybrid structure: as illustrated in Fig.~\ref{fig:teaser}(c), our design integrates a pre-trained LLM layer for global modeling in a hybrid structure. It takes the CNN-encoder-extracted feature as input, projected by a linear layer, and outputs processed tokens to the CNN decoder. Given that the LLM layer is initialized with pre-trained weights and kept frozen, this design substantially reduces the reliance on large-scale data needed for training ViTs from scratch, while introducing only a minimal increase in the number of trainable parameters.
Quantitative results show that this design exhibits strong generalizability and outperforms baseline architectures across various modalities and domains, including ultrasound, dermoscopy, polypscopy, and CT, 
establishing new state-of-the-art (SOTA) results. 

Despite the promising improvement, considering that LLMs are exclusively trained on text data, \textit{what roles does the LLM play during medical image segmentation}? 
Our analysis demonstrates a possibility that LLM can effectively transfer the semantic awareness learned during pretraining to enhance the understanding of visual semantics. Specifically, in segmentation, a clearer separation between the foreground and background can be found in feature activation maps with the help of an LLM layer, leading to a significant reduction in noise. This semantic refinement, likely stemming from the strong semantic processing ability of LLMs, provides more accurate guidance and enhances local-modeling ability of CNN. Moreover, the improvement exhibits strong robustness. Regardless of input modality or structural configuration, this enhancement remains effective, highlighting the adaptability of LLMs in diverse medical segmentation tasks.

In summary, our contributions can be summarized as: 
\begin{itemize}
    \item We prove the feasibility of leveraging LLMs to process visual tokens in segmentation and propose a novel hybrid framework that incorporates a pre-trained LLM layer for global understanding in segmentation. 
    \item Our design enhances baseline performances and achieves SOTA results across multiple medical imaging domains, including 2D modalities such as polypscopy, dermoscopy, and ultrasound, as well as 3D imaging such as CT scans.
    \item We thoroughly analyze the role of LLM during visual processing and show their potential of generalizable semantic understanding which benefits both global and local modeling in medical image segmentation.
\end{itemize}
\section{Method}
In this section, we introduce a novel LLM4Seg framework designed to explore the capabilities of pre-trained LLMs for processing visual tokens, with a focus on segmentation tasks. Unlike conventional approaches, our framework replaces the transformer block in a hybrid pipeline with a frozen LLM layer.

\noindent\textbf{\underline{Hybrid segmentation model recap.}}
    Hybrid approaches leverage the strengths of both CNN and Transformer architectures, combining local feature extraction with global contextual modeling for improved segmentation performance.
    A hybrid model for segmentation tasks generally comprises two main stages: a CNN stage and a Transformer stage. Given an input image $x$, a CNN $\text{Encoder}(\cdot)$ extracts position-aware activations, denoted as $t\in R^{C\times H\times W}$, which capture localized spatial features. These activations are then flattened to $t'\in R^{C\times HW}$ and fed into $\text{Transformer}(\cdot)$ to enhance global contextual understanding. Finally, the processed tokens are reshaped back and decoded to produce the segmentation prediction $\tilde{x}$. The whole process is represented as:
\begin{equation}
t = \text{Encoder}(x; \theta_1), \quad t' = \text{Flatten}(t),
\end{equation}
\begin{equation}
\label{eq:transformer}
\hat{t} = \text{Transformer}(t'; \gamma), \quad \hat{t}' = \text{Reshape}(\hat{t}),
\end{equation}
\begin{equation}
\tilde{x} = \text{Decoder}(\hat{t}'; \theta_2).
\end{equation}

In recent state-of-the-art models~\cite{tang2023mobileutr,tang2024hyspark,chen2021transunet,wenxuan2021transbts}, the parameters $\theta_1, \gamma, \theta_2$ are typically learned from scratch. However, identifying the optimal parameters for transformer layers often requires access to large-scale training datasets, as their capacity for modeling complex relationships is highly dependent on the volume and diversity of the data.

\noindent\textbf{\underline{LLM-boosted hybrid framework.}}
To leverage the effectiveness of LLMs in processing visual semantics, as illustrated in Fig.\ref{fig:teaser}, we propose a simple yet effective hybrid architecture. Specifically, we replace the transformer layer with an LLM layer, augmented by linear projection layers before and after the LLM block. In this design, the original $\text{Transformer}(\cdot)$ in Eq.~(\ref{eq:transformer}) is substituted with the LLM-based formulation in Eq.~(\ref{eq:llm}). This modification leverages the pre-trained capabilities of LLMs for a global semantic understanding while maintaining the compatibility with the existing framework.
\begin{equation}
\label{eq:llm}
    \begin{aligned}
        \hat{t}&=\text{Linear}(t',\gamma_1 ),\\
        \hat{t}&=\text{Transformer}(\hat{t},\lambda_{LLM}), \\
        \hat{t}&=\text{Linear}(\hat{t},\gamma_2).
    \end{aligned}
\end{equation}

By leveraging a {\bf pre-trained and fixed} $\lambda_{LLM}$, we benefit from a stable and reliable initialization that minimizes the need for extensive retraining. This approach preserves the pre-trained knowledge with minimal modifications, allowing us to effectively assess the role of the LLM in the segmentation task while keeping the increase in computational resources minimal. 

\section{Experiments}
\subsection{Setup}
\noindent\textbf{Datasets.} We use breast ultrasound BUSI~\cite{al2020dataset}, thyroid ultrasound TNSCUI~\cite{tnscui}, dermoscopy ISIC~\cite{isic}, and polypscopy Kvasir~\cite{pogorelov2017kvasir} for 2D evaluation, and abdomen CT dataset BTCV~\cite{btcv} for 3D evaluation. We use a 7/3 split on BUSI, TNSCUI, and ISIC for training and validation. Following previous works, both the Kvasir and BTCV datasets are partitioned into their official training and validation sets~\cite{unetr}, with Kvasir additionally splitting a testing set for evaluation~\cite{pogorelov2017kvasir}.

\noindent\textbf{Evaluation metrics and baseline models.}
Following pervious work~\cite{tang2024cmunext,tinyunet,unext,ma2025towards}, we utilize IoU and F1 score for BUSI, TNSCUI, ISIC, and Kvasir, while adopting the Dice for BTCV as ~\cite{unetr,swinunetr,roy2023mednext}. Trainable params (M) and inference FLOPs (G) are also included for comparison.
We select 11 recent SOTA models for comparison, including 2D segmentation networks: U-Net~\cite{ronneberger2015u}, U-Net++~\cite{unetplus}, TransUNet~\cite{chen2021transunet}, UNeXt~\cite{unext}
 MissFormer~\cite{missformer}, UCTransNet~\cite{uctransnet}, UniRepLKNet~\cite{unireplknet} and TinyU-Net~\cite{tinyunet}; 3D segmentation networks: MedNeXt~\cite{roy2023mednext} and 3D UX-Net~\cite{lee3d}.

\noindent\textbf{Implementation details.}
We use U-Net~\cite{ronneberger2015u}, CMUNeXt~\cite{tang2024cmunext}, and nnUNet~\cite{nnunet} as 2D backbones, and MedNeXt~\cite{roy2023mednext}, 3D UX-Net~\cite{lee3d} as 3D backbones. By default, we empirically employ the frozen 15-th layer of LLaMA3.2-1B~\cite{llama3} or 28-th DeepSeek-R1-Distill-Qwen-1.5B~\cite{deepseek} as the LLM layer. We also implement a trainable version for it, denoted as ``{+LLaMA(T)} / +DeepSeek(T)''.
For fair comparison, we 
follow the same parameter settings and data augmentation as prior 2D~\cite{tang2024cmunext,unext} (256$\times$256 as 2D inputs) and 3D~\cite{unetr,swinunetr} (96$\times$96$\times$96 with spacing 1.5$\times$1.5$\times$2.0 mm as voxel inputs) works. To compare with the baseline, we use a randomly initialized transformer with exactly the same structure as the corresponding LLaMA / DeepSeek layer, denoted as ``+Transformer / +\underline{Transformer}''.

\subsection{Quantitative Results}

\noindent\textbf{\underline{Comparison with other models.}} Surprisingly, Table~\ref{tab:msd} shows that with a frozen LLaMA layer, the segmentation performance generally improves across various benchmarks compared to previous methods, achieving new SOTA average results to 80.63\% in average IoU (from the model ``UNet + LLaMA'') and 87.66\% in average F1 (from the model ``CMUNeXt + LLaMA''). 
Notably, compared to other modality datasets, ISIC features larger foreground regions with distinct edges and lower dependence on global context. Despite this, our design still achieves notable improvements, and it achieves superior average segmentation performance while requiring significantly fewer trainable parameters.

\begin{table}[!t]
\caption{Result on BUSI, TNSCUI, ISIC, and Kvasir. Best results are highlighted as \colorbox{g}{\bf \!first\!}, \colorbox{l}{\!second\!} and \colorbox{ll}{\!third\!}. Param: trainable parameters. \label{tab:msd}}
\footnotesize
\centering
\resizebox{1\linewidth}{!}
{
\begin{tabular}{l |rr|  cc  cc  cc cc  cc}
\hline 

\multirow{2}{*}{Method} 
& \multicolumn{2}{c|}{Computation} 
& \multicolumn{2}{c}{BUSI}
& \multicolumn{2}{c}{TNSCUI}
& \multicolumn{2}{c}{ISIC}
& \multicolumn{2}{c}{Kvasir}
&  \multicolumn{2}{c}{Avg}
\\
\cline{2-13}
 & \multicolumn{1}{c}{Param(M)} & \multicolumn{1}{c|}{GFLOPs} & \multicolumn{1}{c}{IoU} & \multicolumn{1}{c}{F1} & \multicolumn{1}{c}{IoU} & \multicolumn{1}{c}{F1} & \multicolumn{1}{c}{IoU} & \multicolumn{1}{c}{F1} & \multicolumn{1}{c}{IoU} & \multicolumn{1}{c}{F1} & \multicolumn{1}{c}{IoU} & \multicolumn{1}{c}{F1} \\
\hline
TinyU-Net  & 0.48 & 1.66 & 66.21 & 75.01 & 74.03 & 82.95 & 81.95 & 88.99 & 86.87 & 92.51 & 77.26 & 84.86 \\ 
UNeXt  & 1.47 & 0.58 & 65.04 & 74.16 & 71.04 & 80.46 & 82.10 & 89.93 & \cellcolor{ll}{88.01} & 92.05 & 76.54 & 84.15 \\ 
UniRepLK  & 5.83 & 9.39 & 65.26 & 73.96  & 67.73 & 77.20 & 81.64 & 88.82 & 85.30 & 91.18 & 74.98 & 82.78 \\
U-Net++  & 26.90 & 37.62 & 69.49 & 78.06 & 76.90 & 85.13 & 82.29 & 89.17 & 86.60 & 92.08 & 78.82 & 86.11 \\
MissFormer  & 35.45  & 7.25 & 63.29 & 73.47 & 68.26 & 76.71 & 80.99 & 88.03 & 86.37 & 91.79 & 74.72 & 82.50
\\
UCTransNet & 66.24  & 32.98 & 70.05 & 78.51 &75.51 & 84.08 & 82.76 & 89.50 & 87.97 & 92.93 & 79.07 & 86.25  \\
TransUnet  & 105.3 & 38.52 & 71.39 & 79.85 & 77.63 &  85.76 &  \cellcolor{ll}{82.95} & \cellcolor{ll}{89.68} & 87.95 & 92.84 & 80.03 & 87.25 \\
\hline
UNet  & 34.52 & 65.56 & 68.79 & 77.00 & 75.99 & 84.24 & 82.28 & 89.19 & 86.01 & 91.49 & 78.26 & 85.48 \\
+LLaMA & +4.19 & +16.64  & \cellcolor{l}{72.95} & \cellcolor{l}{81.46} & \cellcolor{ll}{77.80} & \cellcolor{ll}{85.93} & 82.88 & 89.57 & \cellcolor{l}{88.90} & \cellcolor{l}{93.53} & \cellcolor{g}{\bf80.63} & \cellcolor{l}{87.62} \\
\hline
CMUNeXt & 3.14 & 7.39 & 71.20 & 79.71 & 76.95 & 85.19 & 82.47 & 89.36  & 87.17 & \cellcolor{ll}{92.95} & 79.44 & 86.80  \\
+LLaMA  & +1.05 & +16.03 &  \cellcolor{g}{\bf73.14} & \cellcolor{g}{\bf81.55} & 77.51 & 85.75 & 82.69 & 89.45 & \cellcolor{g}{\bf89.03} & \cellcolor{g}{\bf93.90} & \cellcolor{l}{80.59} & \cellcolor{g}{\bf87.66}  \\
\hline
nnUNet & 7.76 & 12.12 & 72.76 & 80.72 & \cellcolor{l}{79.53} & \cellcolor{l}{87.34} & \cellcolor{l}{83.07} & \cellcolor{l}{89.51} & 84.97 & 91.15 & 80.08 & 87.18 \\
+LLaMA & +2.10 & +16.10 & \cellcolor{ll}{72.90} & \cellcolor{ll}{80.86} & \cellcolor{g}{\bf79.93} & \cellcolor{g}{\bf87.62} & \cellcolor{g}{\bf83.39} & \cellcolor{g}{\bf89.79} & 84.60 & 90.87 & \cellcolor{ll}{80.21} & \cellcolor{ll}{87.29} \\

\hline
\end{tabular}
}
\end{table}

\begin{table}[!t]
\caption{Comparison with baselines. Improvements over baseline are highlighted as \colorbox{r}{$> 2\%$},  \colorbox{o}{$> 1\%$} and \colorbox{lll}{$> 0.1\%$}.  \textbf{val} (bold): top method. Param: trainable parameters. \label{tab:2dbaseline}}
\centering
\resizebox{1\linewidth}{!}
{
\begin{tabular}{c|c|rr| cc cc cc cc |c| c}
\hline 
 \multirow{2}{*}{Methods}
& \multirow{2}{*}{2D} 
& \multicolumn{2}{c|}{Computation} 
& \multicolumn{2}{c}{BUSI}
& \multicolumn{2}{c}{TNSCUI}
& \multicolumn{2}{c}{ISIC}
& \multicolumn{2}{c|}{Kvasir}
& \multirow{2}{*}{3D}
& BTCV
\\
\cline{3-12}\cline{14-14}
 & & \multicolumn{1}{c}{Param(M)} & \multicolumn{1}{c|}{GFLOPs} & \multicolumn{1}{c}{IoU} & \multicolumn{1}{c}{F1} & \multicolumn{1}{c}{IoU} & \multicolumn{1}{c}{F1} & \multicolumn{1}{c}{IoU} & \multicolumn{1}{c}{F1} & \multicolumn{1}{c}{IoU} & \multicolumn{1}{c|}{F1} &  & \multicolumn{1}{c}{Dice} \\
\hline
\multicolumn{1}{l|}{baseline1}&\multirow{4}{*}{{\rotatebox{90}{UNet}}}& 34.52 & 65.56 & 68.79 & 77.00 & 75.99 & 84.24 & 82.28 & 89.19 & 86.01 & 91.49 & \multirow{4}{*}{{\rotatebox{90}{MedNeXt}}} & 82.00 \\

\multicolumn{1}{l|}{+Transformer}& & +65.01 & +16.64 &  \cellcolor{r}{72.68} & \cellcolor{r}{81.00} & \cellcolor{r}{77.81} & \cellcolor{o}{85.87} & \cellcolor{lll}{82.63} & \cellcolor{lll}{89.29} & \cellcolor{r}{87.86} & \cellcolor{o}{92.94}  &  & 81.85 \\

\multicolumn{1}{l|}{+LLaMA(T)} & & +65.01 & +16.64 & \cellcolor{r}{72.63} & \cellcolor{r}{80.94} & \cellcolor{r}{\textbf{77.83}} & \cellcolor{o}{85.84} & \cellcolor{lll}{\textbf{83.01}} & \cellcolor{lll}{89.62} & \cellcolor{r}{88.97} & \cellcolor{r}{93.65} &  & \cellcolor{lll}{82.43} \\

\multicolumn{1}{l|}{+LLaMA } & & +4.19 & +16.64 & \cellcolor{r}{72.95} & \cellcolor{r}{81.46} & \cellcolor{r}{77.80} & \cellcolor{r}{\textbf{85.93}} & \cellcolor{lll}{82.88} & \cellcolor{lll}{89.57} & \cellcolor{r}{88.90} & \cellcolor{r}{93.53}  &  & \cellcolor{lll}{\textbf{82.55}} \\
\hline
\multicolumn{1}{l|}{baseline2} & \multirow{7}{*}{{\rotatebox{90}{CMUNeXt}}} & 3.14 & 7.39 & 71.20 & 79.71 & 76.95 & 85.19 & 82.47 & 89.36  & 87.17 & 92.95 &  \multirow{7}{*}{{\rotatebox{90}{3D UX-Net}}} & 80.78 \\

\multicolumn{1}{l|}{+Transformer} & & +61.87 & +16.03 & \cellcolor{lll}{71.67} & \cellcolor{lll}{80.26} & \cellcolor{lll}{77.11} & \cellcolor{lll}{85.41} & \cellcolor{lll}{82.62} & \cellcolor{lll}{89.43} & \cellcolor{lll}{87.38} & 92.61 &  & \cellcolor{lll}{81.26} \\

\multicolumn{1}{l|}{+LLaMA(T)} & & +61.87 & +16.03 & \cellcolor{r}{72.66} & \cellcolor{o}{81.05} & \cellcolor{lll}{77.54} & \cellcolor{lll}{85.75} & \cellcolor{lll}{82.92} & \cellcolor{lll}{\textbf{89.70}} & \cellcolor{o}{88.49} & \cellcolor{lll}{92.96}  &  & \cellcolor{lll}{81.71} \\

\multicolumn{1}{l|}{+LLaMA } & & +1.05 & +16.03 & \cellcolor{r}{\textbf{73.14}} & \cellcolor{r}{\textbf{81.55}} & \cellcolor{lll}{77.51} & \cellcolor{lll}{85.75} & \cellcolor{lll}{82.69} & \cellcolor{lll}{89.45} & \cellcolor{r}{\textbf{89.03}} & \cellcolor{lll}{\textbf{93.90}} &  & \cellcolor{o}{\textbf{81.84}} \\

\multicolumn{1}{l|}{+\underline{Transformer} } & & +47.58 & +12.38 & \cellcolor{lll}{71.32} & \cellcolor{lll}{79.89} & \cellcolor{lll}{77.28} & \cellcolor{lll}{85.47} & \cellcolor{lll}{82.59} & \cellcolor{lll}{89.41} & \cellcolor{lll}{87.80} & \cellcolor{lll}{93.05} &   & \cellcolor{lll}{81.23} \\

\multicolumn{1}{l|}{+DeepSeek(T)} & & +47.58 & +12.38 & \cellcolor{lll}{72.00} & \cellcolor{lll}{80.53} & \cellcolor{lll}{77.18} & \cellcolor{lll}{85.39} & \cellcolor{lll}{82.48} & \cellcolor{o}{89.30} & \cellcolor{o}{88.55} & \cellcolor{lll}{93.54} &  & \cellcolor{lll}{80.86} \\

\multicolumn{1}{l|}{+DeepSeek} & & +0.78 & +12.38 & \cellcolor{lll}{71.91} & \cellcolor{lll}{80.45} & \cellcolor{lll}{77.39} & \cellcolor{lll}{85.61} & \cellcolor{lll}{82.63} & \cellcolor{lll}{89.45} & \cellcolor{o}{88.69} & \cellcolor{lll}{93.49} &  & \cellcolor{lll}{81.48} \\
\hline

\end{tabular}
}
\end{table}

\noindent\textbf{\underline{Comparison with baselines.}} 
As presented in Table~\ref{tab:2dbaseline}, the line labeled ``+Transformer'' represents the performance obtained by training directly on the visual dataset from scratch, using the same structure and the same number of parameters. In contrast, loading pre-trained weights from LLaMA / DeepSeek consistently improves the performance. These findings indicate that the performance gain is not attributable to an increase in parameters, but rather to the crucial role of pre-trained LLM in long-range modeling, which significantly boosts global semantic representation. Moreover, the improvement is robust across 2D and 3D datasets, showing the strong generalization ability of our findings.

\noindent\textbf{\underline{Computational resources.}} We analyze the number of trainable parameters and GFLOPs, as shown in the ``Computation'' columns in Tab.~\ref{tab:msd} and Tab.~\ref{tab:2dbaseline}. With the LLM layer frozen, our approach achieves the best performance while incurring only a minimal increase in computational cost. The additional trainable parameters stem solely from the projection layers before and after the transformer layer. Specifically, incorporating a LLaMA / DeepSeek layer adds only 4.19M parameters to UNet and 1.05M / 0.78M to CMUNeXt. Inference runtime increases only slightly compared to without inserting LLM layer (from 5.0ms to 5.9ms), demonstrating the method's practicality under real-world constraints.
\section{Understanding the Role of LLM in Segmentation}

Despite the intriguing improvement, given that LLMs have never seen visual data during pre-training, our analysis suggests a possibility that the LLM layer can transfer the semantic knowledge acquired during pre-training to enhance the understanding of visual semantics. Simultaneously, the semantic refinement facilitates local-modeling ability of CNN. 
These indications are supported by: Activation Analysis in Sec.~\ref{sec:an-a}, Statistic Analysis in Sec.~\ref{sec:an-sta}, and Structural Analysis in Sec.~\ref{sec:an-stu}.

\begin{figure}[t!]
    \centering
    \includegraphics[width=0.97\linewidth]{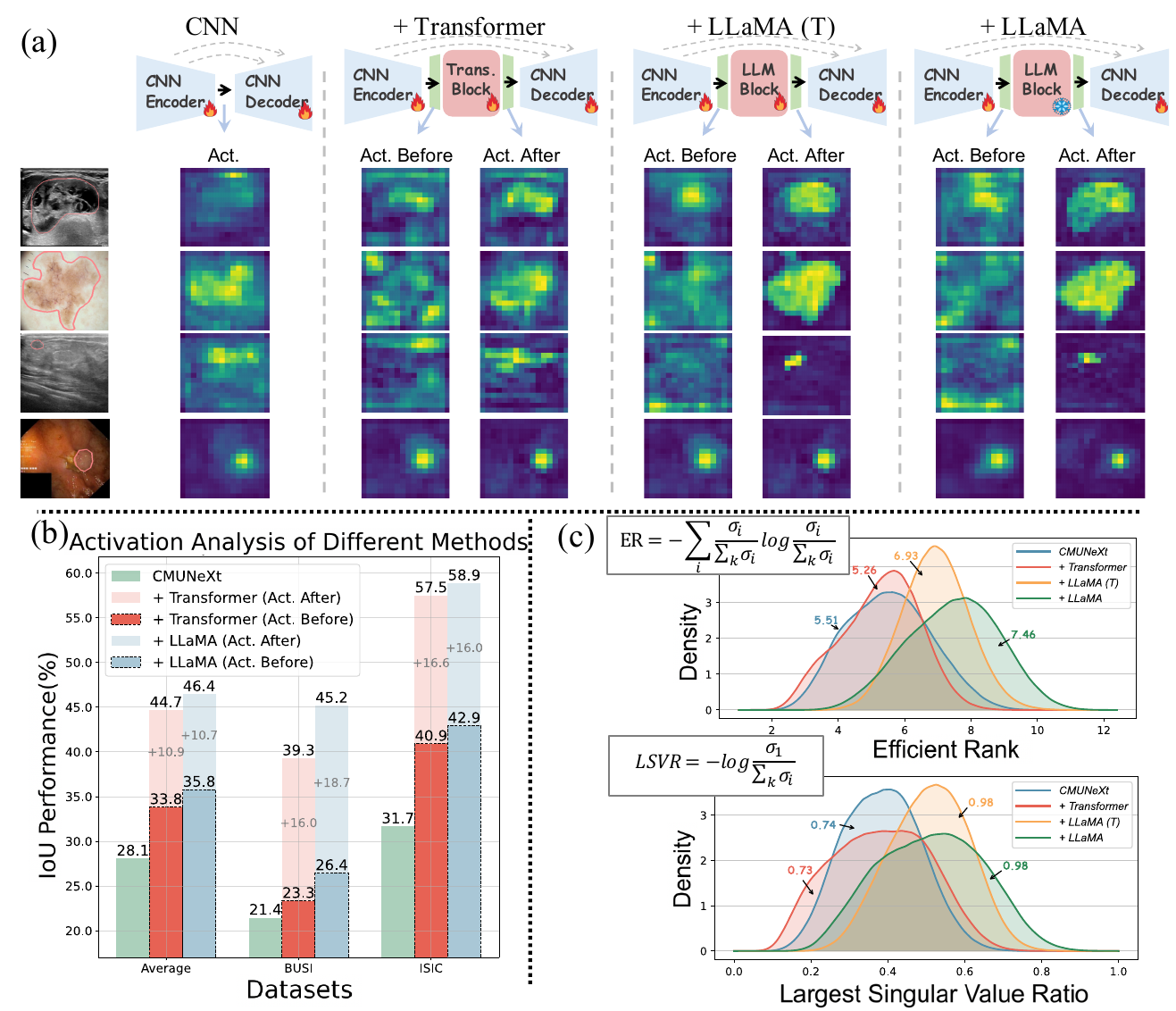}
    \caption{\textbf{(a)} Activation visualization before and after the Transformer/LLaMA layer. \textbf{(b)} Concentration accuracy of activations. The activations are threshold by 0.4 and compared with ground truth segmentation masks to calculate IoU. \textbf{(c)} The distribution of Effective Rank (ER) and Largest Singular Value Ratio (LSVR) of activations extracted from TNSCUI dataset. Average value of each distribution is denoted by $\rightarrow$. }
    \label{fig:vis}
\end{figure}

\subsection{Activation Analysis}
\label{sec:an-a}
\noindent\textbf{\underline{Activation visualization.}} Following ~\cite{pang2023frozen}, we visualize the feature activation maps before and after the Transformer/LLaMA layer, denoted as ``Act. Before'' and ``Act. After''.
As illustrated in Fig.~\ref{fig:vis}(a), a Transformer layer trained from scratch can effectively distinguish the foreground from the background. However, the pre-trained LLaMA layer significantly reduces background noise and produces sharper boundaries, focusing more precisely on lesion regions. This enhanced activation maps underscore the strong transferability of its pre-trained knowledge to novel scenarios, even from text to visual modality, demonstrating the effectiveness of leveraging pre-trained LLaMA layer for visual understanding.

\noindent\textbf{\underline{Activation concentration analysis.}} We further assess the accuracy of activation's concentration by calculating the IoU between ground truth segmentation masks and the highlighted regions in ``Act. Before'' and ``Act. After''. The results, presented in Fig.~\ref{fig:vis}(b), provide a quantitative measure of how effectively the model focuses on relevant foreground regions. 

Considering ``Act. Before'' (darker-colored columns), the concentration accuracy is improved compared to activations from CMUNeXt. Initializing the LLaMA weights further enhances the average IoU score, underscoring LLaMA's role in guiding CNN encoding. Additionally, ``Act. After'' (lighter-colored columns) shows an even greater concentration accuracy compared to those before, reaching its peak with ``+ LLaMA'' design. This result highlights LLaMA's high-level semantic refinement, filtering out background noise and generating more precise activations, which in turn provide reliable information for the CNN decoder.

\subsection{Statistic Analysis}
\label{sec:an-sta}
Following previous works~\cite{roy2007effective,chen2023understanding,parkhi2012cats,chen2019transferability,xue2022investigating}, we analyze the singular value spectrum of the feature space before decoding. Specifically, we perform channel-wise singular value decomposition (SVD) on the activations $\hat{t}'\in R^{C\times H\times W}$ before decoding, obtaining $C$ singular matrix $\Sigma = [\sigma_1, \sigma_2, ..., \sigma_k]\in R^{H\times W}$, where each $\sigma_i$ represents a singular value.

We then calculate two metrics for all singular values: (1) Effective Rank (ER)~\cite{roy2007effective}: the entropy of the singular values, normalized by their sum. Higher ER suggests that the feature space encompasses a greater number of dimensions, thereby more effectively capturing the underlying structure of the data. (2) Largest Singular Value Ratio (LSVR)~\cite{chen2019transferability,chen2023understanding} represents the ratio of the largest singular value to the other singular values, calculated on a negative logarithmic scale. A smaller LSVR indicates greater dominance of the largest singular value, which might overshadow the representation ability from other singular values and damage the model's representation ability.
As shown in Fig. \ref{fig:vis}(c), the pre-trained LLaMA layer outperforms the Transformer layer and the baseline CMUNeXt in terms of both average ER and LSVR, with the frozen LLaMA layer achieving the highest ER. This suggests a broader range of singular values of the features after LLaMA, indicating the potentially stronger representation capacity in the feature space. These results validate our suggestion that LLaMA can facilitate visual representation learning in medical image segmentation tasks.

\subsection{Structural Analysis}
\label{sec:an-stu}
\begin{figure}[h!]
\centering
\includegraphics[width=\textwidth]{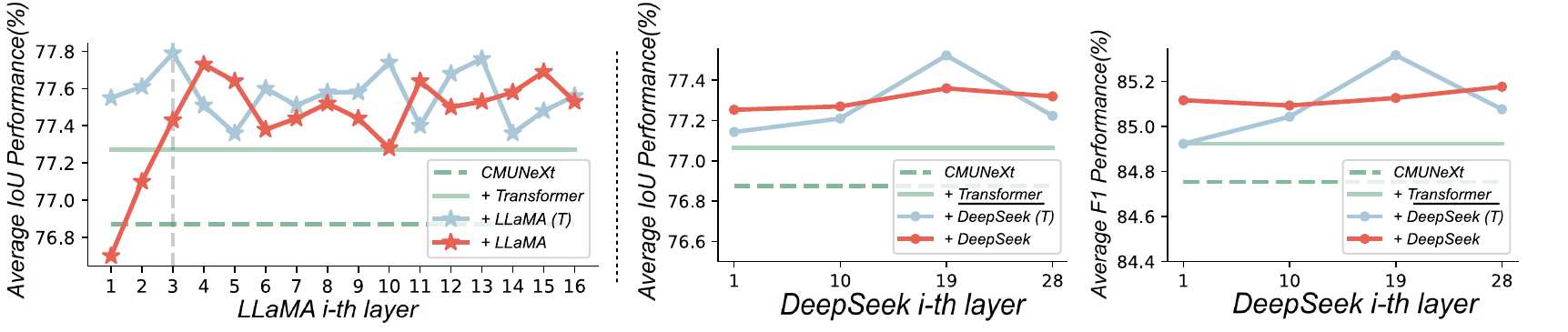}
\caption{Impact of different LLM Transformer layers on segmentation performance.} \label{fig:layers}
\end{figure}

For LLM structure, Fig.~\ref{fig:layers}(Left) reveals that the improvement remains consistent across various layer configurations of LLaMA, provided that the selected layer is sufficiently deep to capture comprehensive semantics for understanding (at least after 3-th layer for LLaMA3.2-1B). Similarly, Fig.~\ref{fig:layers}(Right) demonstrates that inserting a DeepSeek-R1 layer also benefits segmentation with noticeable performance gains when using deeper layers. These results further validate the generalizability of LLM's knowlesge, emphasizing that the effectiveness in medical segmentation is robust to LLM structure and layer selection.

\section{Conclusion and Discussion}
Our findings reveal a novel and unexpected generalization capability of LLMs: their semantic awareness in medical image segmentation. We propose to integrate a frozen pre-trained LLM layer into a CNN architecture, demonstrating improved segmentation performance across various imaging modalities. Because the encoder is trainable, it learns to project features into the input space of the frozen LLM layer, effectively tapping into the semantic priors learned from large-scale language pretraining. Thorough analysis further validates the effectiveness of this design, highlighting the potential of LLMs in visual processing.

These insights open new avenues for multi-modal synergies between language models and vision tasks and can be potentially extended beyond segmentation to challenges like classification, detection, and anomaly identification in medical imaging. However, as shown in our structural analysis, performance fluctuations exist, suggesting room for improvement. Future work may explore more robust adaptation strategies or diverse LLM architectures to enhance stability, optimize performance, and broaden applicability.

\begin{credits}
\subsubsection{\ackname} Supported by Natural Science Foundation of China under Grant 62271465, Suzhou Basic Research Program under Grant SYG202338.

\subsubsection{\discintname}
The authors have no competing interests to declare that are relevant to the content of this article.
\end{credits}

%
%
%
\bibliographystyle{splncs04}
\bibliography{ref}
%




\end{document}